# NEXT-GENERATION PERCEPTION SYSTEM FOR AUTOMATED DEFECTS DETECTION IN COMPOSITE LAMINATES VIA POLARIZED COMPUTATIONAL IMAGING

Yuqi Ding[1], Jinwei Ye[1], Corina Barbalata[2], James Oubre[3], Chandler Lemoine[2], Jacob Agostinho[2] and Genevieve Palardy[2]

[1]Division of Computer Science and Engineering, Louisiana State University
3325 Patrick F. Taylor Hall

[2]Department of Mechanical & Industrial Engineering, Louisiana State University
3261 Patrick F. Taylor Hall

[3]Division of Electrical and Computer Engineering, Louisiana State University
3325 Patrick F. Taylor Hall

Baton Rouge, LA 70803

## ABSTRACT

Finishing operations on large-scale composite components like wind turbine blades, including trimming and sanding, often require multiple workers and part repositioning. In the composites manufacturing industry, automation of such processes is challenging, as manufactured part geometry may be inconsistent and task completion is based on human judgment and experience. Implementing a mobile, collaborative robotic system capable of performing finishing tasks in dynamic and uncertain environments would improve quality and lower manufacturing costs. To complete the given tasks, the collaborative robotic team must properly understand the environment and detect irregularities in the manufactured parts. In this paper, we describe the initial implementation and demonstration of a polarized computational imaging system to identify defects in composite laminates. As the polarimetric images are highly relevant to the surface micro-geometry, they can be used to detect surface defects that are not visible in conventional color images. The proposed vision system successfully identifies defect types and surface characteristics (e.g., pinholes, voids, scratches, resin flash) for different glass fiber and carbon fiber laminates.

## 1. INTRODUCTION

Finishing operations on large-scale composite components, such as wind turbine blades, often require multiple workers and part repositioning. Moreover, these tasks, including trimming and sanding, are tedious for humans, leading to physical discomfort and repetitive strain injuries. In the composites manufacturing industry, automation of such processes is challenging, as manufactured part geometry may be inconsistent and task completion is based on human judgment and experience. Implementing a mobile, collaborative robotic system capable of performing finishing tasks in dynamic and uncertain environments would improve quality and lower manufacturing costs. To complete the given tasks, the collaborative robotic team must properly understand the environment and detect irregularities in the manufactured parts. Laser-based systems (e.g., line laser triangulation, overhead laser template projector) have traditionally been used for part inspection [1-3]. However, they present constraints regarding working distance, angle between laser and imager, and material reflectivity, which would pose a significant challenge for a team of mobile robots performing finishing tasks. While visual characteristics of composite laminates present challenges for state-of-the-art imaging systems, the proposed polarized computational imaging system will address this limitation.

Traditional computer vision algorithms only consider the visual characteristics of composite laminates through the intensity of the image. A traditional computer vision approach to detect defects in carbon fiber/epoxy laminates (dimensions 20.5 cm x 22.0 cm) is presented in Figure 1. Starting from a color image, a grayscale transform is performed (Figure 1a) and a gaussian blur is applied to the image (Figure 1b) to remove the noise [4]. The next step is to get the x and y gradients of the blurred image. These gradients show when there is a change in both the x and y directions and can be used to detect sudden changes in the image [5]. The gradients are then combined through normalization and the defects start to become more apparent. After normalization, the image is blurred again to combine the separate small, detected defects into one region. Next, the gradients are taken again to refine the shape of the defect area as seen in Figure 5e. Morphological operations are then used to fill in the gaps and remove unsubstantial areas that have been detected [6].

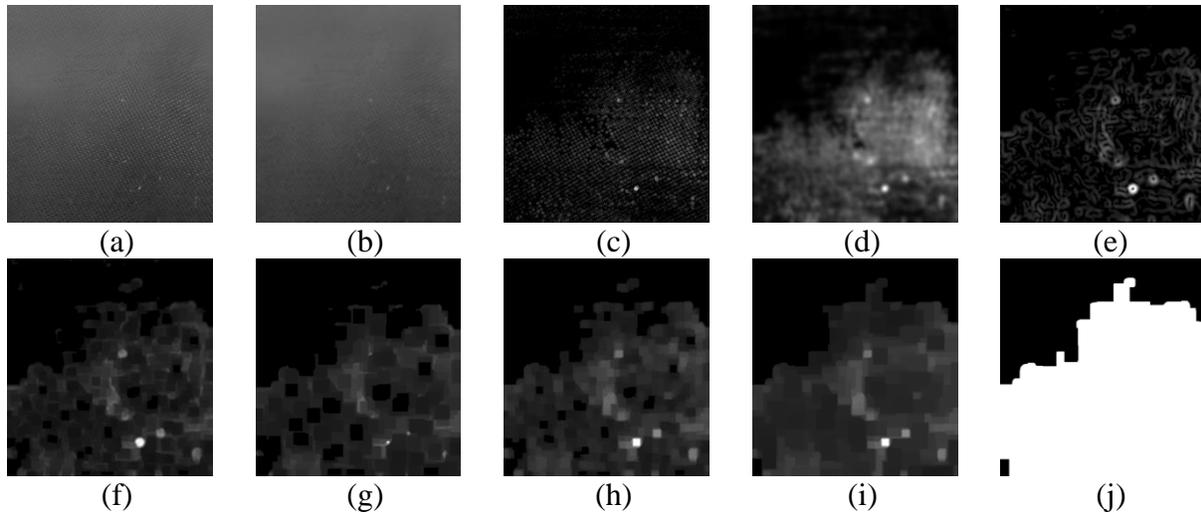

Figure 1. Pipeline of traditional computer vision approaches applied to a carbon fiber/epoxy laminate containing several pinholes: (a) grayscale image; (b) blurred image; (c) normalization of gradients; (d) normalized image blurred; (e) gradient of blurred normalized image; (f) morphological closing; (g) morphological erosion; (h) morphological dilation; (i) second morphological closing; and (j) binary threshold of defected area.

The output of this traditional approach can be seen in Figure 2, where Figure 2a is the original RGB image, Figure 2b represents the ground truth manually labeled, while Figure 2c represents the automated detection of the method described above. As can be seen, the obtained results show rough defect detection. However, fine defects (such as individual pinholes) are difficult to identify. Furthermore, this visual inspection using RGB images is challenging due to the specular reflection and low contrast.

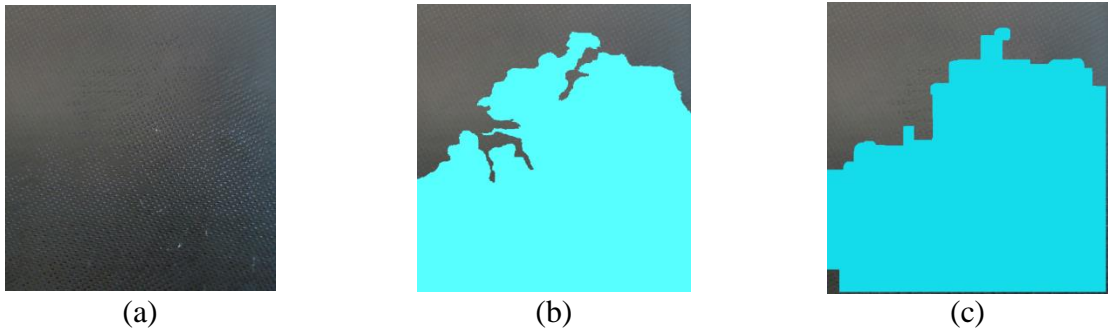

Figure 2. Output of the traditional computer vision approach: (a) input RGB image; (b) manually labeled defects (pinholes); and (c) traditional computer vision approach.

To improve the inspection performance, we propose a novel computational imaging method based on polarization. In this paper, we will discuss the initial implementation and demonstration of this computer vision system to identify defects and surface characteristics in composite laminates. As the polarization images are highly related to the micro-geometry of the material, it can be used to detect surface defects which are not visible through traditional approaches, as shown in Figure 2. Using novel sensor technology, we use a polarization camera, capturing the polarization images in one shot. With those images, we first compute the Stokes parameters [7] that describe the polarization state of light. We then analyze the polarization state with two representations: Degree of Polarization (DoP) and Angle of Polarization (AoP) [8], as pinholes, voids, dry spots, or scratches on surfaces become more distinguishable in these images. Finally, we use the DoP and AoP images for deficiency detection, which helps improve the detection rate. In the experiment, we validate our polarization imaging system with carbon fiber (CF) and glass fiber (GF) laminates. The results show our method can successfully detect different types of defects or surface characteristics.

## 2. METHODOLOGY

### 2.1. Materials and Manufacturing Methods

Several types of composite materials were used to compare the performance of imaging techniques for various features and defects. Table 1 summarizes the materials and manufacturing approaches used for all specimens. Vacuum-assisted resin infusion (VARI) specimens (GF and CF) were used for the traditional computer vision approach summarized in Figures 1 and 2. All other samples were used for the proposed polarimetric computational imaging system to test its limitations with a wider range of defects and materials.

Table 1. Materials and manufacturing methods used in this study.

| Sample name | Reinforcement / Matrix | Layup | Manufacturing method |
|---|---|---|---|
| VARI_GF/epoxy | Saertex 830 gsm Stitched Biaxial Glass Fiber / Epoxy (System 4500) [a] | [(±45)]$_3$* | Vacuum-assisted resin infusion (VARI) |
| VARI_CF/epoxy | Twill Toray T-300 245gsm Carbon Fiber / Epoxy (System 4500) [a] | [(0/90)]$_3$* | VARI |
| UDGF/PP | Unidirectional Glass Fiber / Polypropylene IE 6030 Polystrand$^{TM}$ prepreg [b] | [0]$_8$ | Compression molding with heated press [9] |
| LGF/PP | Complēt® Long Glass Fiber / Polypropylene pellets [c] | Random | Compression molding with heated press |
| UDCF/epoxy | Unidirectional IM7 Carbon Fiber / Epoxy Cycom® 5320 prepreg [d] | [(0/90)$_3$]s* | Out of autoclave (OOA) vacuum bagging and oven heating [10] |
| PWCF/epoxy | Plain weave T650 CF / Epoxy Cycom® 5320 prepreg [d] | [(0/90)$_3$]s* | OOA vacuum bagging and oven heating [10] |

[a] Purchased from FiberGlast; [b] Provided by Avient (formerly PolyOne); [c] Provided by PlastiComp; [d] Cytec Solvay Group, provided by The Boeing Company.
* (±45) and (0/90) denote a single fabric ply.

## 2.2. Polarization Imaging

### 2.2.1. Polarization of Light

Polarization is a fundamental light property that describes the oscillation of the electric wave of the light. The unpolarized light, for example sunlight, usually vibrates at random directions that are perpendicular to the propagation direction. Generally, the unpolarized light becomes polarized light after passing through a polarizer. That is because the polarizer allows the light with certain polarization direction to pass, while other directions are blocked.

The unpolarized light can also be polarized by surface reflection. Specifically, the polarization state of light can be decomposed in two directions: one is perpendicular to the incident plane of reflection, called the s-component, and the other is parallel to the incident plane of reflection, called the p-component. Since these two components are not reflected in the same ratio, the reflected light becomes polarized as one polarization direction becomes dominant over the other, as shown in Figure 3. The Fresnel equations [11] quantify the ratios of reflection with respect to the surface's index of refraction. For this reason, the polarization images are highly sensitive to surface micro-structure.

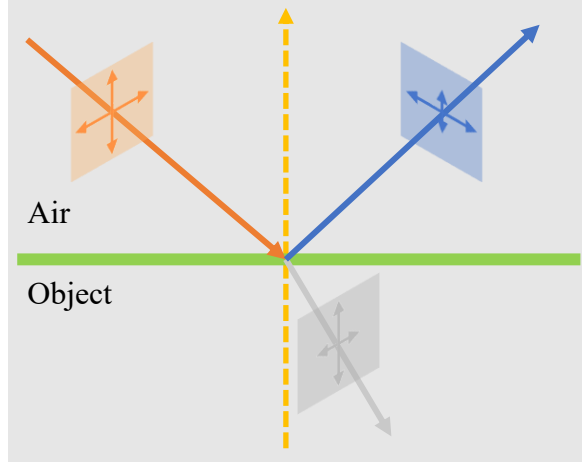

Figure 3. Illustration on how the polarization state of light is changed after reflection or transmission.

Human eyes and traditional optical cameras do not have the polarization sensitivity and cannot directly measure the polarization state of light. In order to measure and analyze the polarization property of light, various polarizers are commonly used. In the next section, we describe the polarization imaging techniques in detail.

*2.2.2. Polarization Imaging*

A straightforward approach to capture polarization images is to place a linear polarizer in front of a traditional camera and take a set of images with the linear polarizer rotated at different angles. The linear polarizer only allows the light whose polarization direction is aligned with its angle to pass through. In this way, one measures the light intensity of various polarization directions. Minimally, one needs to measure under four polarization angles: 0°, 45°, 90°, and 135°. By combining the four measurements, the polarization state of light that enters the camera can be estimated. However, this straightforward approach needs several measurements by rotating the linear polarizer and cannot be used on a moving robotic platform.

In this paper, we leverage new sensing technology and adopt a polarization camera that can measures the polarization state of light in a single shot. Specifically, the model of the polarization camera we use is Lucid Vision Labs Phoenix [12]. It uses a Sony IMX250MZR sensor that adopts the wire grid polarization principle. As shown in Figure 4, the sensor has on-chip linear polarizers at $0°$, $45°$, $90°$ and $135°$ that are located on top of the 3.45 $\mu m$ pixels with Bayer pattern. This allows the camera to capture the four directional polarization images in one shot. The resolution of output image is 2448 × 2048 pixels. This sensor is in global shutter exposure mode and has a frame rate of 22 frames per second.

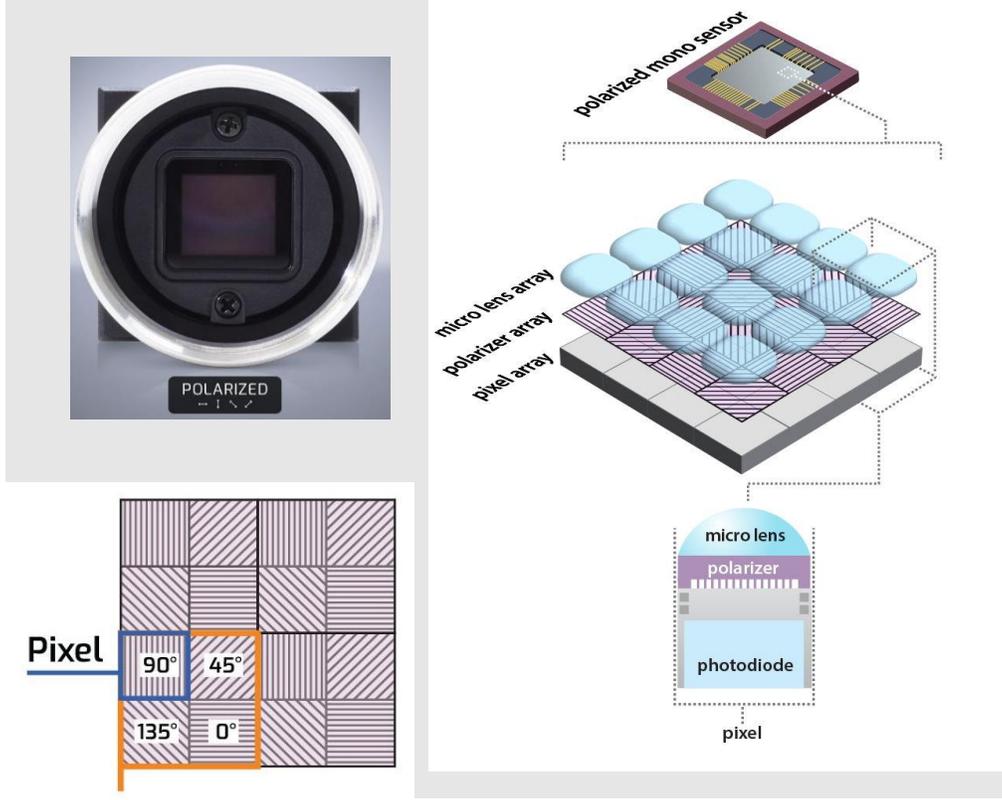

Figure 4. Illustration of structure of the polarization sensor.

### 2.2.3. Stokes Parameters Measurements

Once we have the four directional polarization images, we use them to compute the Stokes parameters for each pixel. The Stokes parameters [7] are a set of real values that describe the polarization state of the light. It contains four elements: $S_0$, $S_1$, $S_2$, and $S_3$, where $S_0$ represents the irradiance of the light, $S_1$ represents the horizontal or vertical polarization state, $S_2$ represents the $45°$ or $135°$ polarization state, and $S_3$ represents the amount of left or right circular polarization. The direction of polarization state represented by the Stokes parameters are relative to a reference axis. Here, we use the x-axis of the image pixel coordinate as the reference axis.

The relationship between the image intensity and the Stokes parameters is shown in Equation [1]:

$$I(\varphi, \theta) = 0.5 * (S_0 + S_1 \cos 2\theta + S_2 \cos \varphi \sin 2\theta + S_3 \sin \varphi \sin 2\theta) \quad [1]$$

where, $\varphi$ is the retardation angle and $\theta$ is the polarization angle. In this paper, we only consider linear polarization. We therefore assume $S_3 = 0$ and $\varphi = 0$.

As we have the intensities of four directional polarization images $I(0°)$, $I(90°)$, $I(45°)$, and $I(135°)$, we can compute the Stokes parameters using the following equations:

$$S_0 = I(0°) + I(90°)$$
$$S_1 = I(0°) - I(90°) \quad [2]$$
$$S_2 = I(45°) - I(135°)$$

where, $I(0°)$, $I(45°)$, $I(90°)$ and $I(135°)$ are the intensity of image under corresponding polarization angle.

### 2.2.4. Polarization State Analysis

To analyze the polarization state of light, we adopt two representations for polarization: the Degree of Polarization (DoP) and Angle of Polarization (AoP) [8]. DoP describes the portion of the polarized light. For instance, the DoP of unpolarized light is 0, whereas the DoP of fully polarized light is 1. DoP ($P$) can be computed using the Stokes parameters as:

$$P = \sqrt{S_1^2 + S_2^2 + S_3^2}/S_0 \quad [3]$$

AoP represents the angle of linear polarization. It indicates the angle between the plane of polarization and the plane of reference. AoP ($\emptyset$) can be computed using Stokes parameters as:

$$\emptyset = 0.5 * \tan^{-1}(S_2/S_1) \quad [4]$$

The AoP value is in the range of [0°, 180°]. Both DoP and AoP of the reflected light are highly relevant to the surface geometry and material properties. Therefore, various types of surface defects, such as pinholes, voids, dry spots, or scratches become more visible in these images.

## 3. RESULTS AND DISCUSSION

### 3.1. Experimental Setting

In the experiment, we use the polarization camera to capture the materials in the uncontrolled environment, which means the light source is unpolarized. The distance between the camera and the material is near 20 *cm*. We choose a 12 *mm* focal length Kowa lens. For the baseline part, we use a Cannon 60D DSLR camera to capture the color images. The resolution of color images is $5184 \times 3456$ and the focal length is 50 *mm*. All computations are performed on a desktop computer with Intel i7 processor, 64 GB memory. Our image processing method is written in Matlab.

### 3.2. Polarization Image versus Color Image

We first capture the polarization images through the Lucid polarization camera. We then decode the raw images to four directional polarization images. We use the decoded images to compute $S_0$, $S_1$ and $S_2$ by applying Equation [2]. To build the baseline, we also capture the color images through DSLR camera. Figures 5 and 6 show a comparison of the cropped images of glass fiber and carbon fiber specimens in different modes. In these images, the size of the composite samples is 38.1 mm x 50.8 mm (1.5" x 2").

In Figure 5, the color images of GF/PP samples are near white and therefore, it is difficult to extract the structure information. The color images only contain the spectrum information. In the polarization images represented by the Stokes parameters ($S_0$, $S_1$, and $S_2$), we can see that the structural information is more clearly revealed. The Stokes parameters can uncover detailed structure information and defects (i.e., voids, fiber orientation). This is because these structural changes in surface geometry, although hardly visible in the spectral domain due to limited camera resolution, may cause drastic variations in the polarization state of reflected light, which makes them highly distinguishable in polarization images. In Figure 6, the color images of CF/epoxy samples cannot identify obvious defects. In the polarization images represented by the Stokes parameters ($S_0$, $S_1$, and $S_2$), the defects and surface characteristics (i.e., pinholes, scratches) are visible with higher contrast.

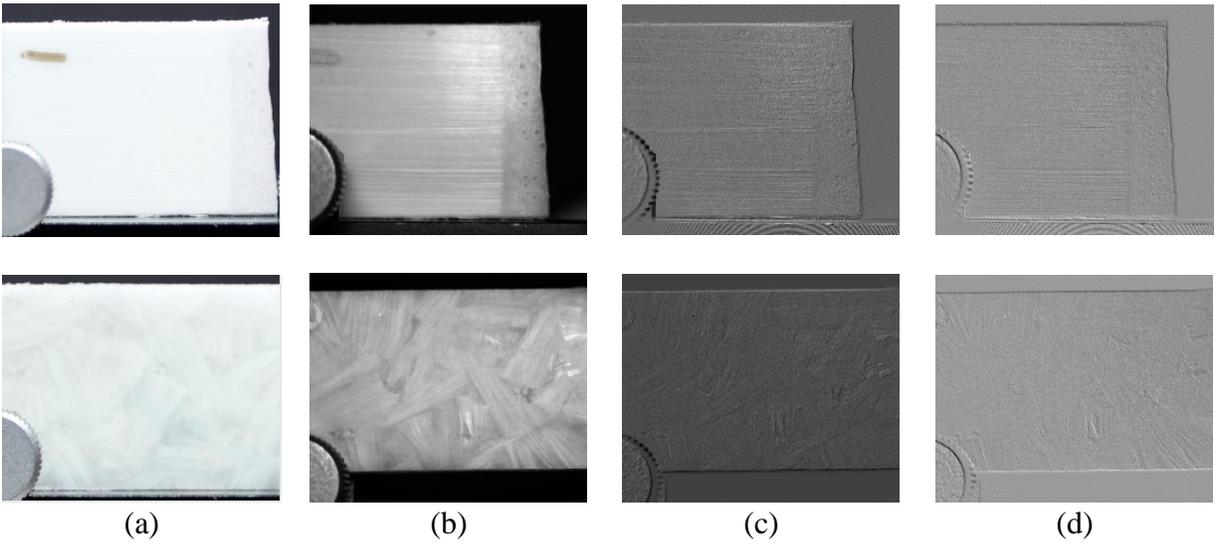

(a) (b) (c) (d)

Figure 5. UDGF/PP (top row) and LGF/PP (bottom row) samples. (a) Color Image; (b) $S_0$; (c) $S_1$; and (d) $S_2$.

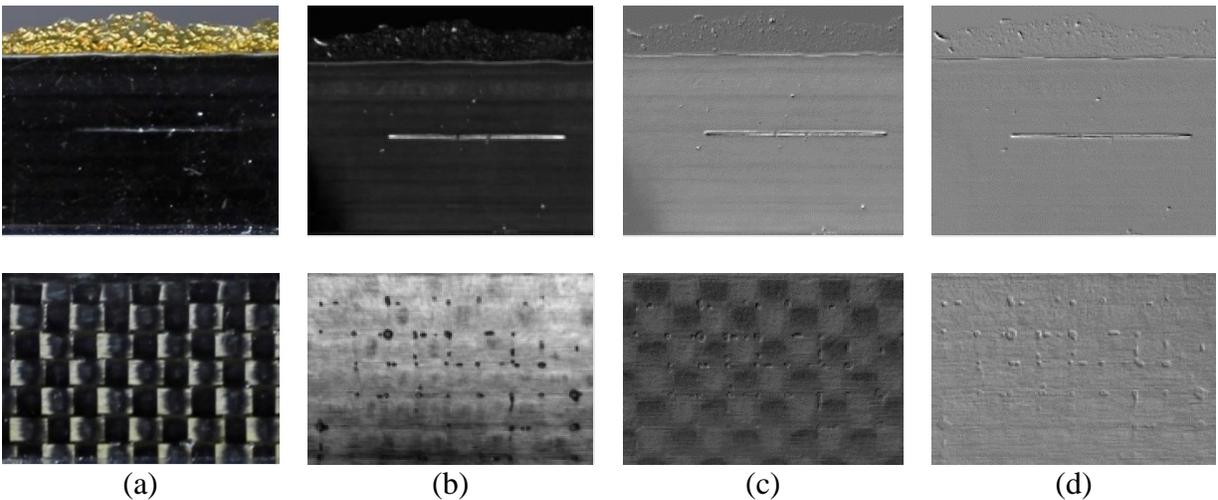

(a) (b) (c) (d)

Figure 6. UDCF/epoxy (top row) and PWCF/epoxy (bottom row) samples. (a) Color Image; (b) $S_0$; (c) $S_1$; and (d) $S_2$.

## 3.3. Defects Detection

For more effective defect detection, we compute DoP and AoP for the material samples using the Stokes parameters by applying Equations [3] and [4]. Figure 7 shows the results of the GF/PP samples. Since diffuse reflection dominates the reflection of the samples, the DoP is relatively low and the AoP is noisy. However, these two polarization representations still show some useful structure information, such as voids and fiber orientation. Figure 8 shows the results of the CF/epoxy samples. For those laminates, specular reflection has a strong effect, which means the DoP is relatively high. The defects are more clearly observed in the DoP images, compared to the GF/PP samples: scratches and resin flash in UDCF/epoxy, and pinholes and resin flash in PWCF/epoxy.

We also apply classical image processing methods on images shown in Figures 5 and 6 to see if it benefits defects detection. We use the Sobel filter on these images to obtain the edges [13]. Figures 9 and 10 show the edge detection results of the GF/PP samples and the CF/epoxy samples, respectively. We can see that the edge detection results on the color images (column (a) in Figures 9 and 10) barely reveal visual defects (voids, pinholes, scratches, resin flash). For GF/PP specimens, $S_0$ results (column (b) in Figure 9) display defects and surface characteristics the most clearly. For CF/epoxy, the edge detection results on $S_2$ (column (d) in Figure 10) show defects are more distinguishable. While not all features shown in Figures 9 and 10 correspond to surface defects or characteristics, we note that this method allows detection of much finer defects compared to traditional computer vision approaches.

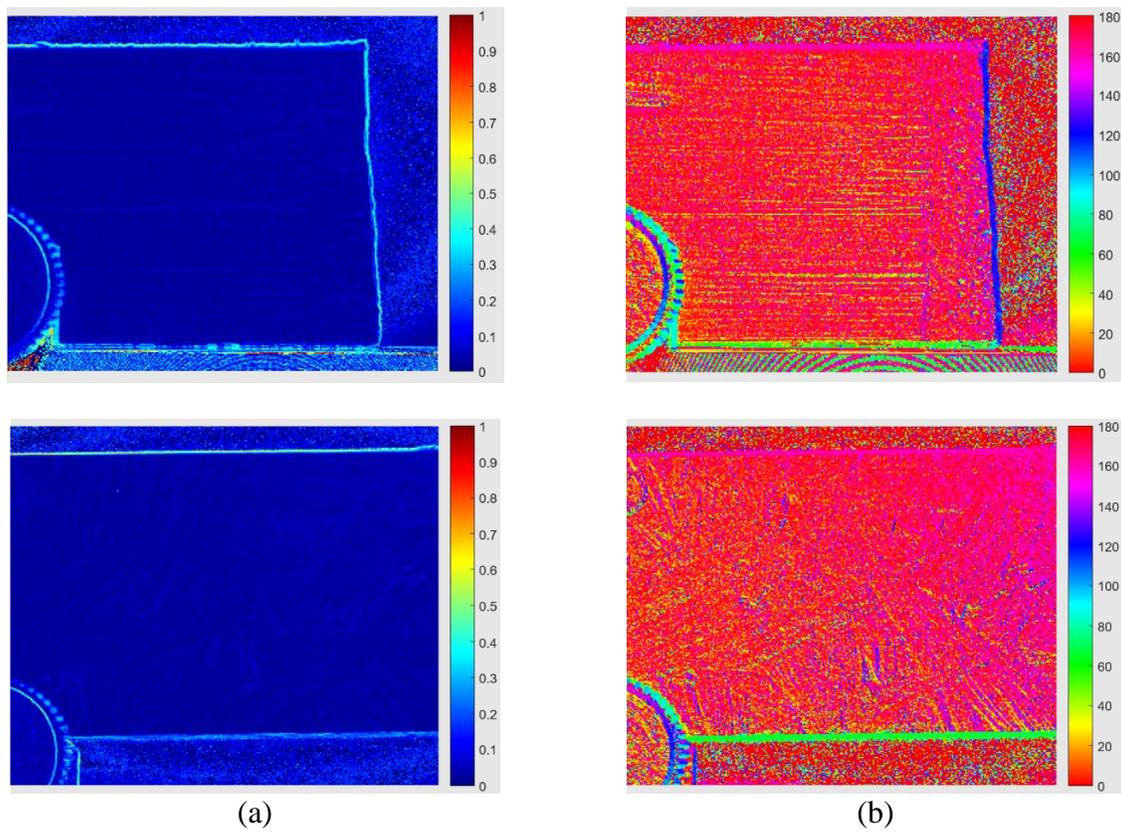

(a)        (b)

Figure 7. UDGF/PP (top row) and LGF/PP (bottom row) samples. (a) DoP; and (b) AoP.

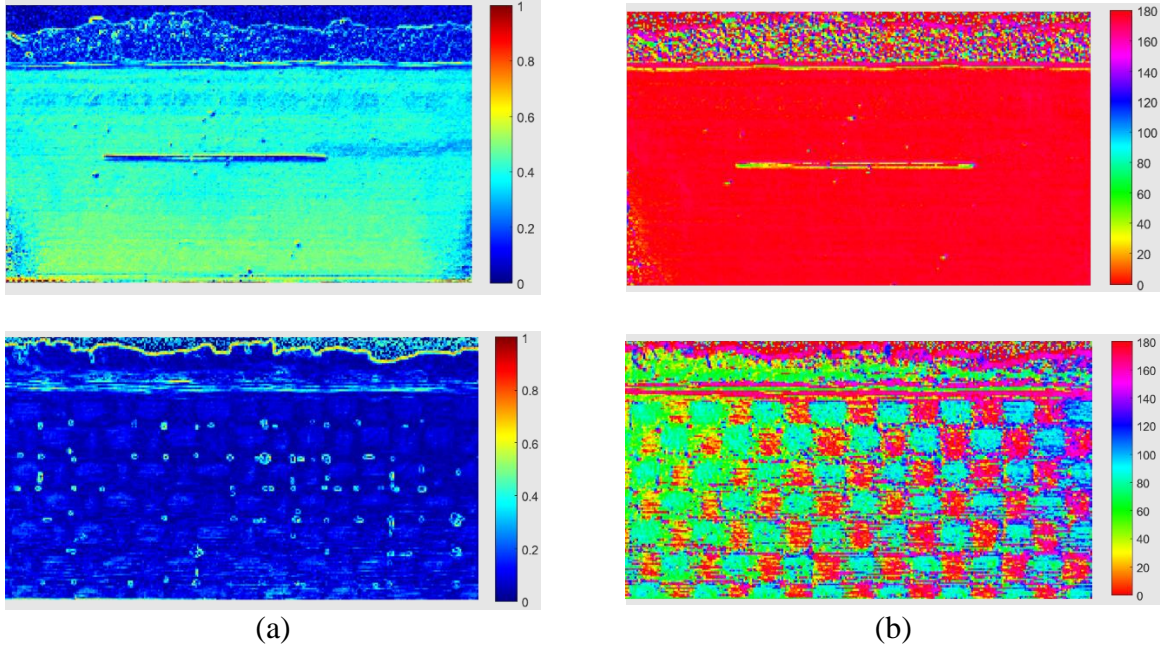

Figure 8. UDCF/epoxy (top row) and PWCF/epoxy (bottom row) samples. (a) DoP; and (b) AoP.

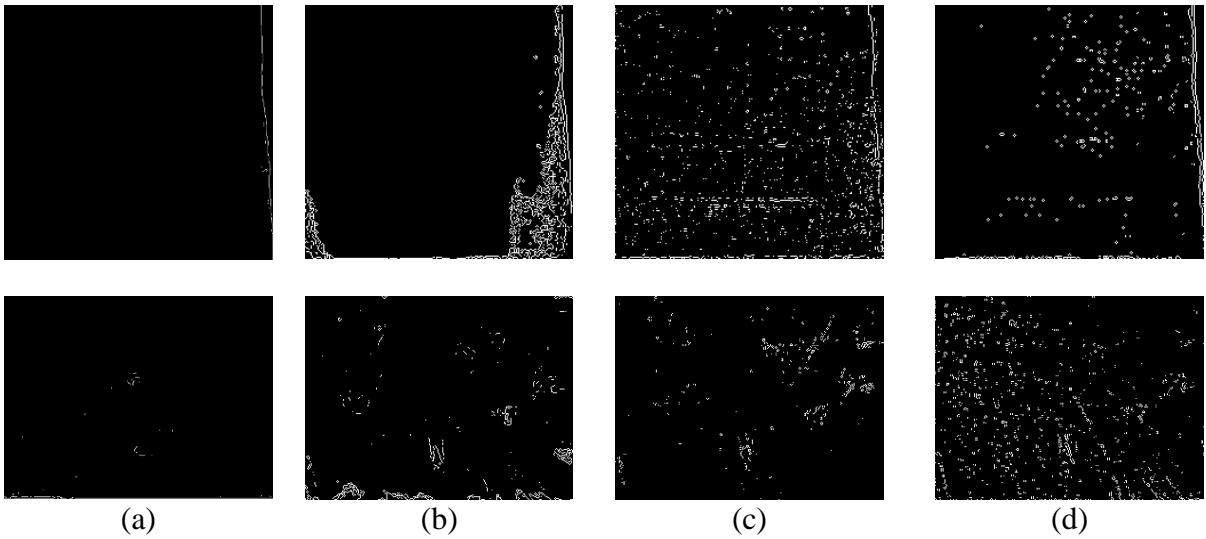

Figure 9. Edge detection result of UDGF/PP (top row) and LGF/PP (bottom row) samples. (a) Color Image; (b) $S_0$; (c) $S_1$; and (d) $S_2$.

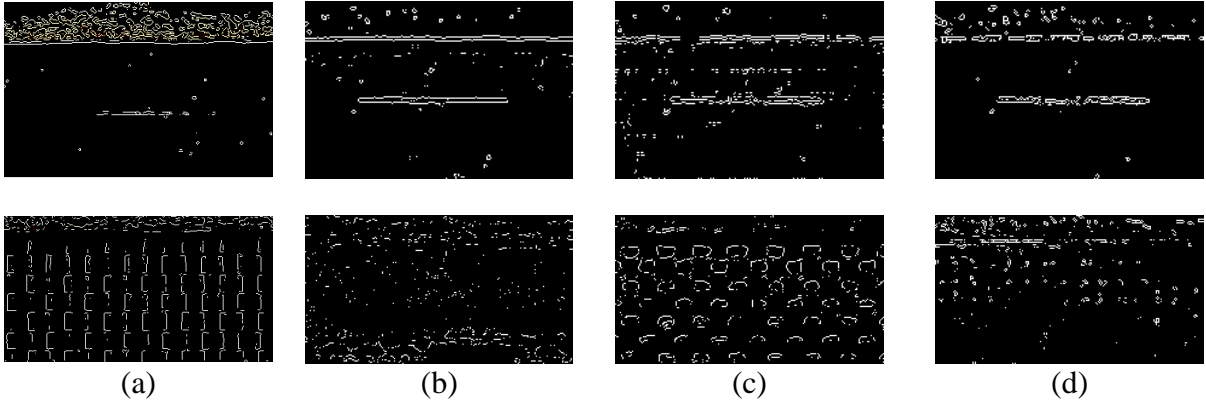

Figure 10. Edge detection result of UDCF/epoxy (top row) and PWCF/epoxy (bottom row) samples. (a) Color Image; (b) $S_0$; (c) $S_1$; and (d) $S_2$.

## 4. CONCLUSIONS

Based on polarization computational imaging, a defect detection method was performed for several types of laminates, reinforced with glass fiber or carbon fiber. Through polarimetric images, we can directly measure the Stokes parameters of the reflected light in one shot. The polarization state can be represented as DoP and AoP through the Stokes parameters. Compared with the traditional vision system that captures color images, the polarimetric images can offer richer information to uncover microstructural information. Defects (i.e., pinholes, voids, scratches) and surface characteristics (i.e., fiber orientation and pattern) have higher contrast in the polarimetric images than color images. This study showed that Stokes parameters, DoP and AoP can be used for the polarization image technique to improve defects and surface characteristics detection based on the type of material (glass fiber vs carbon fiber). Those outcomes will be leveraged as a starting point for integration into a mobile robotic system and further automation of defects detection.

## 5. ACKNOWLEDGMENTS

The authors would like to acknowledge financial support from the following sources: National Science Foundation Award Number 2024795, "NRI: FND: Collaborative Mobile Manufacturing in Uncertain Scenarios", and Louisiana Board of Regents under the Research Competitiveness Subprogram (contract number LEQSF (2018-2022)-RD-A-05).